\documentclass{article} %
\usepackage{iclr2024_conference, times}

\usepackage{amsmath,amsfonts,bm}

\def\eqref#1{equation~\ref{#1}}

\def\1{\bm{1}}

\DeclareMathAlphabet{\mathsfit}{\encodingdefault}{\sfdefault}{m}{sl}
\SetMathAlphabet{\mathsfit}{bold}{\encodingdefault}{\sfdefault}{bx}{n}

\usepackage{microtype}
\usepackage{graphicx}
\usepackage{booktabs} %

\usepackage{hyperref}
\usepackage{tabularx}
\newcolumntype{Y}{>{\tiny\arraybackslash}X}

\usepackage{amsmath}
\usepackage{amssymb}
\usepackage{mathtools}
\usepackage{amsthm}
\usepackage{multirow}
\usepackage{floatrow}
\usepackage{bm}
\usepackage{enumitem}
\usepackage{floatrow}
\usepackage{wrapfig}
\usepackage{titletoc}
\usepackage{placeins}
\usepackage{tikz}
\usepackage[export]{adjustbox}

\usepackage{xcolor}
\usepackage[capitalize]{cleveref}  %

\theoremstyle{plain}

\theoremstyle{definition}

\theoremstyle{remark}

\usepackage{subfig}

\renewcommand{\vec}[1]{\mathbf{#1}}
\newcommand{\mat}[1]{\vec{#1}}

\newcommand{\weightsout}{\bm{\psi}}
\newcommand{\gatingout}{\bm{\chi}}
\newcommand{\norm}[1]{\left\lVert{#1}\right\rVert_2}

\newcommand{\rotatingdim}{N}
\newcommand{\featuredim}{D}
\newcommand{\batchdim}{B}
\newcommand{\inputdim}{\featuredim_{\text{in}}}
\newcommand{\outputdim}{\featuredim_{\text{out}}}

\newcommand{\kerneldim}{K}
\newcommand{\heightdim}{H}
\newcommand{\widthdim}{W}

\newcommand{\inputlayer}{\vec{z}_\text{in}}
\newcommand{\outputlayer}{\vec{z}_\text{out}}

\newcommand{\mbind}{\vec{m}_{\text{bind}}}

\newcommand{\mbo}{MBO}
\newcommand{\mboc}{\mbo$_c$}
\newcommand{\mboi}{\mbo$_i$}

\usepackage[textsize=tiny]{todonotes}
\title{Binding Dynamics in Rotating Features}

\author{Sindy Löwe$^1$ ~~~~ Francesco Locatello$^2$ ~~~~ Max Welling$^1$ \\
        $^1$AMLab, University of Amsterdam ~~$^2$Institute of Science and Technology Austria (ISTA)
}

\begin{document}

\maketitle

\begin{abstract}
In human cognition, the binding problem describes the open question of how the brain flexibly integrates diverse information into cohesive object representations. Analogously, in machine learning, there is a pursuit for models capable of strong generalization and reasoning by learning object-centric representations in an unsupervised manner. Drawing from neuroscientific theories, Rotating Features learn such representations by introducing vector-valued features that encapsulate object characteristics in their magnitudes and object affiliation in their orientations. The ``$\bm{\chi}$-binding'' mechanism, embedded in every layer of the architecture, has been shown to be crucial, but remains poorly understood. In this paper, we propose an alternative ``cosine binding'' mechanism, which explicitly computes the alignment between features and adjusts weights accordingly, and we show that it achieves equivalent performance. This allows us to draw direct connections to self-attention and biological neural processes, and to shed light on the fundamental dynamics for object-centric representations to emerge in Rotating Features.
\end{abstract}

{\let\thefootnote\relax\footnotetext{
Correspondence to \href{mailto:loewe.sindy@gmail.com}{loewe.sindy@gmail.com}
}
\newcommand{\thefootnote}{\arabic{footnote}}

\section{Introduction}
Recognizing and conceptualizing objects lies at the heart of human cognition, allowing us to interact with and understand our environment and to apply these insights to novel situations \citep{wertheimer1922untersuchungen,koffka1935principles,kohler1967gestalt}. However, the binding problem remains a significant challenge. It questions how our brain integrates diverse information to create cohesive object representations within a network of relatively fixed connections. Answering this question would not only enhance our understanding of human cognition, but could also lead to machine learning models with similarly strong generalization and reasoning abilities as our brains \citep{greff2020binding}.

With the development of Rotating Features \citep{lowe2022complex,lowe2023rotating}, a new direction of research has emerged for studying the binding problem in machine learning. Taking inspiration from neuroscience, Rotating Features are vector-valued features, which represent object characteristics in their magnitudes and object affiliation in their orientation. They employ a binding mechanism, ``$\gatingout$-binding'', which ensures that features with similar orientations are processed together, while connections between features with dissimilar orientations are weakened. While this binding mechanism has been shown to be crucial for Rotating Features to overcome the binding problem and to create object-centric representations, it remains poorly understood. For one, the dynamic described above has only been proven to hold for directly opposing features (i.e. features with cosine similarity of $-1$) and under limiting assumptions \citep{reichert2013neuronal,lowe2022complex}. Additionally, its implementation, which applies the same weights to the input features and their magnitudes before combining the results, is difficult to interpret. As a result, it is unclear how this binding mechanism processes features under more realistic scenarios; and what kind of dynamics are truly necessary to enable Rotating Features to bind object features effectively.

In this paper, we introduce an alternative, ``cosine binding'' mechanism for Rotating Features to improve our understanding of the dynamics required to make them work. By explicitly computing the alignment between features and weighing inputs based on this alignment, it follows a series of intuitive steps. As it achieves a similar performance to the $\gatingout$-binding mechanism, this allows us to learn more about the dynamics necessary to enable Rotating Features to learn object-centric representations. Additionally, it allows us to draw connections to dynamics observed in biological neurons and to the implementation of self-attention \citep{vaswani2017attention}. Overall, we believe that our proposed binding mechanism enhances our understanding of Rotating Features and can thus help practitioners to improve them.

\section{Background: Rotating Features}

Rotating Features are vector-valued features, which learn to represent object characteristics in their magnitudes and object affiliation in their orientations. To implement this, a standard $\featuredim$-dimensional feature vector $\mathbf{z}_{\texttt{standard}} \in \mathbb{R}^{\featuredim}$ is expanded to a feature matrix $\mathbf{z}_{\texttt{rotating}} \in \mathbb{R}^{\rotatingdim \times \featuredim}$ containing $\rotatingdim$-dimensional Rotating Features. To ensure that features with similar orientations are processed together, the model implements a binding mechanism. Given the input $\inputlayer \in \mathbb{R}^{\rotatingdim \times \inputdim}$ to a layer with $\inputdim$ input and $\outputdim$ output feature dimensions, it applies the weights $\mathbf{w} \in \mathbb{R}^{\inputdim \times \outputdim}$ to both the input features and their magnitudes\footnote{Omitting biases for simplicity.}, before combining the results:
\begin{align}
    \weightsout &= f_{\mathbf{w}}(\inputlayer) ~~~ \in \mathbb{R}^{\rotatingdim \times \outputdim} \label{eq:1} \\
    \gatingout &= f_{\mathbf{w}}(\norm{\inputlayer}) ~~~ \in \mathbb{R}^{\outputdim} \label{eq:2} \\
    \mbind &= 0.5 \cdot \norm{\weightsout} + 0.5 \cdot \gatingout ~~~ \in \mathbb{R}^{\outputdim} \label{eq:3}
\intertext{
Here, $\norm{\cdot}$ denotes the L2-norm over the rotation dimension $\rotatingdim$. The resulting values $\mbind$ are rescaled using batch normalization \citep{ioffe2015batch} and a ReLU non-linearity before they are used as the magnitudes of the layer's output $\outputlayer \in \mathbb{R}^{\rotatingdim \times \outputdim}$:}
    \outputlayer &= \frac{\weightsout}{\norm{\weightsout}} \cdot \text{ReLU}(\text{BatchNorm}(\mbind)) \label{eq:nonlinearity}
\end{align}
To build a network with Rotating Features, this formulation is integrated into every layer of an autoencoding architecture. Then, the network is trained to reconstruct images in the output's magnitudes and the network's object discovery performance is evaluated based on the output's orientation values.

As highlighted by \citet{reichert2013neuronal}, the mechanism described in \cref{eq:1,eq:2,eq:3}, which we term $\gatingout$-binding, results in features with similar orientations being processed together, while connections between features with dissimilar orientations are suppressed. This is crucial to enable the network to learn to create object-centric features, as highlighted by ablation studies by \citet{lowe2022complex,lowe2023rotating}. However, this dynamic has only been proven under limiting assumptions, and it is unclear how the above equations behave under more realistic scenarios.

\section{Cosine Binding in Rotating Features}

We develop the cosine binding mechanism that follows a series of intuitive steps to improve our understanding of the dynamics necessary to enable Rotating Features to learn object-centric representations. In this section, we describe this binding mechanism in more detail, before drawing connections to biological neural processes and to self-attention \citep{vaswani2017attention}.

\subsection{Implementation}
In the proposed cosine binding mechanism, we utilize the cosine similarity between the input and intermediate output features to determine the weighing of each input feature. To accomplish this, we first apply the weights $\mat{W} \in \mathbb{R}^{\inputdim \times \outputdim}$ to the inputs $\vec{x} \in \mathbb{R}^{\batchdim \times \rotatingdim \times \inputdim}$ to compute an intermediate output $\vec{y} \in \mathbb{R}^{\batchdim \times \rotatingdim \times \outputdim}$: 
\begin{align}
    \vec{y}_{bnj} &= \sum_i \mat{W}_{ij} \vec{x}_{bni}
\end{align}
Here, $\batchdim$, $\rotatingdim$, $\inputdim$, and $\outputdim$ refer to the batch, rotation, input, and output dimensions, indexed with $b$, $n$, $i$, and $j$, respectively. Next, we calculate the rescaled cosine similarity $\vec{c} \in [0, 1]^{\batchdim \times \inputdim \times \outputdim}$ over the rotation dimension $\rotatingdim$ of the input $\vec{x}$ and the intermediate output $\vec{y}$:
\begin{align}
    \vec{c}_{bij} &= S_C(\vec{x}_{bi}, \vec{y}_{bj}) * 0.5 + 0.5
\end{align}
We adjust the cosine similarities $S_C(\vec{x}_{bi}, \vec{y}_{bj}) \in [-1, 1]^{\batchdim \times \inputdim \times \outputdim}$ to lie within the range $[0,1]$, as preliminary experiments show that this yields better results. Overall, this gives a measure of alignment for the features involved in each connection between an input and an output neuron. We use this measure to adjust the layer's weights, resulting in a new weight $\mat{W}' \in \mathbb{R}^{\batchdim \times \inputdim \times \outputdim}$ for each neuronal connection, per input sample:
\begin{align}\label{eq:rescaled_weight}
    \vec{W}'_{bij} = \vec{c}_{bij} \vec{W}_{ij}
\end{align}
We apply these adjusted weights to the initial input and add the biases $\mathbf{b} \in \mathbb{R}^{\rotatingdim \times \outputdim}$:
\begin{align}
    \vec{z}_{bnj} &= \sum_i \mat{W}'_{bij} \vec{x}_{bni} + \mathbf{b}_{nj}
\end{align}
We then use a non-linearity similar to \cref{eq:nonlinearity} on the resulting values $\vec{z} \in \mathbb{R}^{\batchdim \times \rotatingdim \times \outputdim}$, which rescales the magnitude of each Rotating Feature in $\vec{z}$ without altering its orientation:
\begin{align}
    \outputlayer &= \frac{\vec{z}}{\norm{\vec{z}}} \text{ReLU}(\text{BatchNorm}(\norm{\vec{z}})
\end{align}
This creates the layer's final output $\outputlayer \in \mathbb{R}^{\batchdim \times \rotatingdim \times \outputdim}$. To implement cosine binding in a convolutional layer, we follow the same procedure while using a differently structured weight matrix.

Compared to $\gatingout$-binding, cosine binding follows a series of intuitive steps, and thus provides useful insights into the dynamics required for Rotating Features to learn object-centric representations. While $\gatingout$-binding has been proven to mask inputs opposite in orientation to the output, given certain conditions \citep{reichert2013neuronal,lowe2022complex}, cosine binding shows this behavior trivially, with no assumptions necessary. Moreover, understanding how $\gatingout$-binding functions in more realistic scenarios is challenging. The dynamics of cosine binding, on the other hand, are simple to comprehend in varied situations. Inputs are weighed differently in the final output based on their alignment with the intermediate outputs they have generated. Perfectly aligned inputs maintain their original weight, and lesser aligned inputs are reweighed based on their cosine similarity to the output. This explicit implementation improves our understanding of the binding dynamics needed for Rotating Features to work, and enables us to draw connections to biological neuronal processes and to self-attention, which we discuss in the subsequent sections.

\subsection{Connection to Neuroscience}\label{sec:neuroscience}

Rotating Features were developed based on mechanisms theorized to underlie object representations in the brain. Particularly, they were inspired by the temporal correlation hypothesis \citep{singer1995visual, singer2009distributed}, which suggests that neurons convey two message types via their spiking patterns: (1) the firing rate of a neuron encodes feature presence, and (2) the synchronization between neurons' spiking indicates whether the represented features should be processed jointly. Hence, simultaneous firing by neurons represents that their corresponding features belong to the same object. This representational format is embedded in the oscillating firing patterns of neurons, also known as brain waves. Rotating Features abstract these temporal dynamics to develop a representational format that utilizes the same two message types: (1) neurons represent feature presence via their magnitudes and (2) object affiliation via their orientation. Thus, by sharing similar orientations, neurons represent that their respective features belong to the same underlying object. 

\begin{figure}[t]
    \centering
    \includegraphics[height=5.1cm]{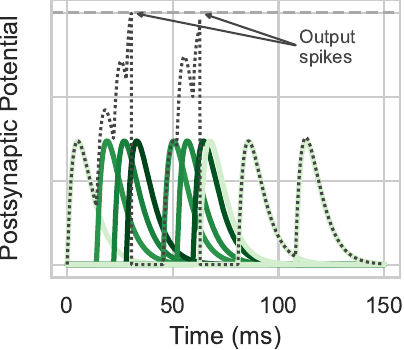} \hspace{0.05\linewidth}
    \includegraphics[height=5.1cm]{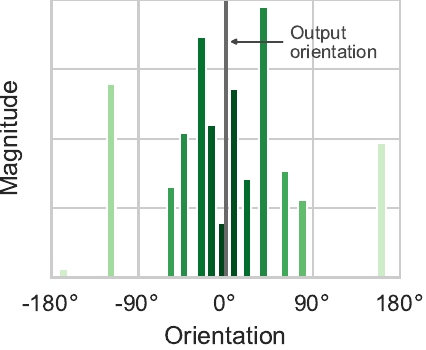}
    \caption{Coincidence detection in biological neurons (\textbf{left}) compared to cosine binding in Rotating Features (\textbf{right}). In biological neurons, incoming spikes are integrated temporally. When the resulting membrane potential (dotted line) reaches threshold (dashed horizontal line), an output spike is emitted, and the membrane potential is reset. Synchronously arriving spikes (dark green) trigger an output spike, while incoming spikes that arrive immediately after an output spike or occur individually (light green) have no impact on the output. In cosine binding, groups of aligned input features result in an intermediate output of similar orientation (gray line, centered at zero). Then, inputs are weighed based on their alignment with this intermediate output (highlighted by different shades of green) leading to the masking of unaligned features. In both setups, synchronization regulates signal transmission. Aligned inputs jointly generate the output and are thus processed together, while unaligned inputs are masked to minimize their influence on this output.
    }
    \label{fig:CoincidenceDetection}
\end{figure}

The cosine binding mechanism enhances the link between Rotating Features and their biological counterpart by drawing parallels to the ``integrate-and-fire'' model of biological neurons \citep{sherrington1906integrative,eccles1957physiology}. Under this model, neurons aggregate the synaptic potential of incoming spikes over time until a threshold is reached, and an outgoing spike is emitted. Given the high sensitivity of biological neurons to temporal differences between incoming spikes, and the quick reset of their membrane potential in the absence of new spikes, they can also be interpreted as coincidence detectors. This suggests that neurons only spike if incoming spikes happen within a tight temporal window. \citet{konig1996integrator} provide extensive arguments supporting the notion that coincidence detection is a plausible mechanism for neuronal communication -- it is physiologically viable on a millisecond timescale (3-5ms), processes information quickly, filters out noisy inputs, and enables complex computations by using the temporal dimension to encode information.

Under both the coincidence detection model for biological neurons and the proposed cosine binding mechanism for Rotating Features, the alignment between input signals is crucial. In biological neurons (\cref{fig:CoincidenceDetection} -- left), incoming spikes only generate an output spike when they occur synchronously, while desynchronized spikes have no effect on the output. Similarly, in the proposed cosine binding mechanism (\cref{fig:CoincidenceDetection} -- right), aligned inputs produce an output of similar orientation, leading to the masking of misaligned inputs. In both setups, synchronization regulates signal transmission. Aligned inputs are processed together, while misaligned inputs are prevented from impacting the output. As a result, the network can process features independently of one another by creating distinct groups of aligned inputs. This naturally leads to the emergence of object-centric representations, as the network clusters the features of different objects for separate processing.

\subsection{Connection to Self-Attention}
Both the cosine binding mechanism and self-attention \citep{vaswani2017attention} leverage the notion of alignment between input features to determine their impact. Cosine binding calculates the cosine similarity between the inputs and intermediate outputs, and weigh inputs based on this score. Meanwhile, self-attention determines alignment between different tokens by computing the inner product between their queries and keys. The resulting attention weights enable aligned tokens to be processed together, and mask misaligned tokens. However, this approach is less precise than the cosine binding mechanism. For one, self-attention computes alignment between tokens, i.e. feature vectors, not between individual features. Additionally, it works with linear projections (query, key and value) rather than the inputs directly. Finally, the inner product used to calculate the attention weights does not solely rely on orientation, but is also influenced by the magnitude of the tokens. Overall, although both mechanisms leverage the concept of alignment for determining input influence, cosine binding leverages this dynamic to create features that represent object affiliation via their orientation.

\section{Experiments}
To improve our understanding of Rotating Features and how they learn object-centric representations, we introduce the cosine binding mechanism. In this section, we examine how this mechanism compares to its predecessor, $\gatingout$-binding, in terms of its performance and computational efficiency.

\paragraph{Object Discovery Performance} To compare the two binding mechanism, we integrate them in the same Rotating Features architecture and apply the resulting models to two real-world datasets, Pascal VOC and FoodSeg103 (see \cref{app:implementation} for more details). Then, we evaluate their object discovery performance by comparing the segmentation masks created by clustering their output orientations to the ground truth. The results are shown in \cref{tab:performance}. We find that the baseline of the Rotating Features model without binding mechanism (\textit{no binding}) performs at random, achieving a similar performance to \textit{block masks} which partition the images into regular rectangles. The \textit{$\gatingout$-binding} and \textit{cosine binding} mechanisms, on the other hand, create a meaningful object separation and perform very similarly, with the $\gatingout$-binding mechanism performing better on Pascal VOC and the cosine binding mechanism performing better on FoodSeg103.

\begin{table}
    \centering
    \caption{Object discovery performance on the Pascal VOC and FoodSeg datasets (mean $\pm$ sem across 5 seeds). The \mboi{} and \mboc{} scores indicate a good separation of objects on these real-world datasets for both the $\gatingout$-binding and cosine binding mechanisms. Results for $\gatingout$-binding and block masks taken from \citet{lowe2023rotating}.
    }
    \label{tab:performance}
    \resizebox{\linewidth}{!}{%
    \begin{tabularx}{\linewidth}{l@{\hskip 30pt}l@{\hskip 40pt}rY@{\hskip 10pt}rY}
        \toprule
        Dataset & Model & \multicolumn{2}{l}{\mboi ~$\uparrow$} & \multicolumn{2}{l}{\mboc ~$\uparrow$} \\
        \midrule
        Pascal VOC  &     Block masks             & 0.247 &              & 0.259              \\
                    &     No binding              & 0.223 & $\pm$ 0.015  & 0.258 & $\pm$ 0.017 \\ 
                    &     $\gatingout$-binding       & 0.407 & $\pm$ 0.001  & 0.460 & $\pm$ 0.001 \\  %
                    &     cosine binding       & 0.397 & $\pm$ 0.006  & 0.449 & $\pm$ 0.005 \\ %
        \midrule
        FoodSeg103  &   Block masks             & --    &       & 0.296            \\
                    &   No binding              & --    &       & 0.275 & $\pm$ 0.078 \\ 
                    &   $\gatingout$-binding       & --    &       & 0.484 & $\pm$ 0.002 \\
                    &   cosine binding       & --    &       & 0.522 & $\pm$ 0.005 \\  %
        \bottomrule    
    \end{tabularx} 
    }
\end{table}

\paragraph{Object Separation}
\Cref{fig:objects_on_circle} illustrates a difference in the object separation learned by the $\gatingout$- and cosine binding mechanisms. Here, we used the same Rotating Features architecture with the respective binding mechanisms on the 2Shapes and 3Shapes datasets \citep{lowe2022complex}, using a rotation dimension of $\rotatingdim=2$. This allows us to visualize the orientations that the models assign to different objects. From this, we see that $\gatingout$-binding results in maximally separated orientations: when two objects are present, this mechanism assigns them opposing orientations; and when there are three objects, they are distributed roughly $120^{\circ}$ apart. In comparison, cosine binding results in substantially closer orientation grouping for features belonging to different objects. Nonetheless, both mechanisms create accurate object segmentations in their predictions.

\paragraph{Runtime and Memory Efficiency}
The cosine binding mechanism involves calculating the cosine similarity between every pair of input/output features, followed by rescaling the connecting weights. This leads to increased memory requirements, particularly in a convolutional layer. Conventionally, in two-dimensional convolutional layers, the weight matrix $\vec{W}$ is of size $[\inputdim, \outputdim, \kerneldim, \kerneldim]$, where $\inputdim, \outputdim$ are the input and output feature dimensions, and $\kerneldim$ is the size of the convolutional kernel. The matrix $\vec{W}'$ (\cref{eq:rescaled_weight}) that is rescaled depending on the alignment between every input/output feature pair, on the other hand, is of size $[\batchdim, \inputdim, \outputdim, \heightdim, \widthdim, \kerneldim, \kerneldim]$ where $\batchdim$ is the batch size and $\heightdim, \widthdim$ denote the feature map's height and width. While the number of learnable parameters stays the same, within our implementation, this amounts to an increase in size by a factor of 1024 to 16384 per weight matrix. This results in the model with $\gatingout$-binding requiring 240MB of GPU storage while cosine binding needs 8400MB (see \cref{app:efficiency} for details). Consequently, the cosine binding mechanism also takes longer to run -- approximately 380ms for a single forward-backward pass through the model -- compared to the 70ms required by the $\gatingout$-binding mechanism.
These high runtime requirements of cosine binding point towards an exciting research direction utilizing spiking neural networks. Here, the same functionality could be emulated with no computational overhead by implementing neurons as coincidence detectors (\cref{sec:neuroscience}). \citet{zheng2023gust} have made some promising strides in this direction.

\newcommand{\mycolumnsep}{1em}
\newcommand{\smallpicsize}{2cm}

\begin{figure}[t]
    \centering
    \begin{tabular}{r@{\hskip \mycolumnsep}cc@{\hskip 20pt}cc}
        & \multicolumn{2}{c}{$\gatingout$-binding} & \multicolumn{2}{c}{cosine binding} \vspace{0.3em} \\               
        \adjustbox{valign=m}{Input} &
        \adjustbox{valign=m}{\includegraphics[width=\smallpicsize]{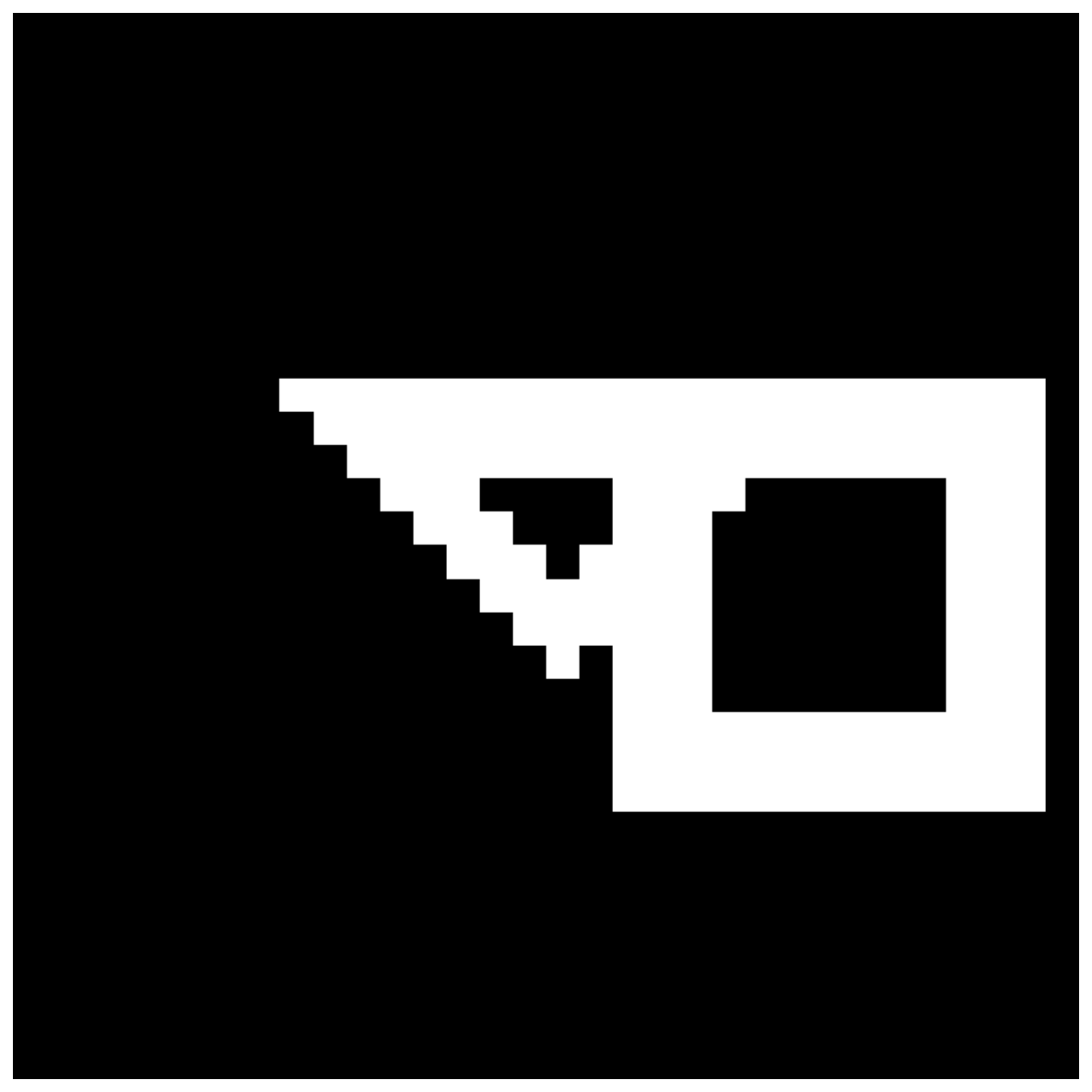}} & 
        \adjustbox{valign=m}{\includegraphics[width=\smallpicsize]{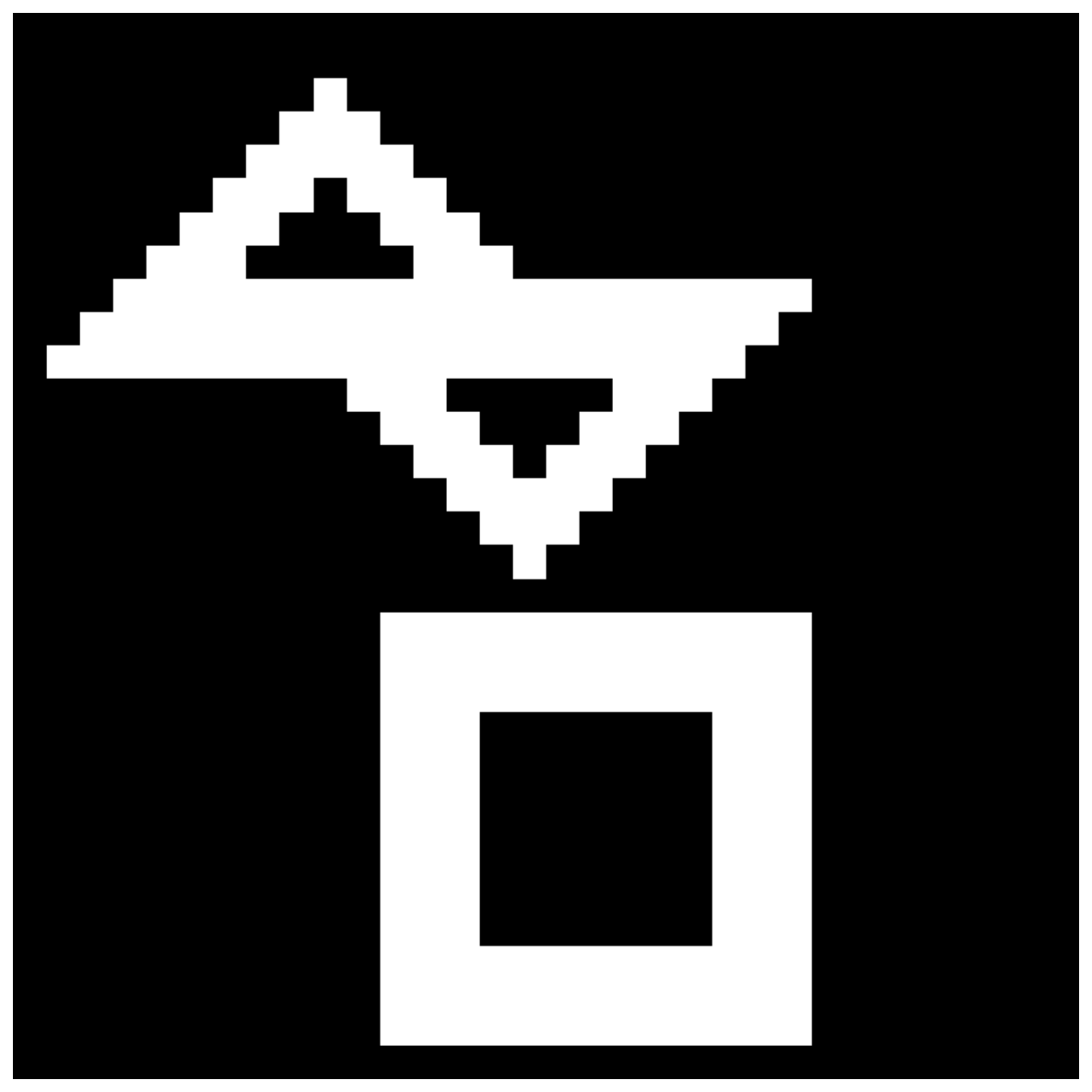}} & 
        \adjustbox{valign=m}{\includegraphics[width=\smallpicsize]{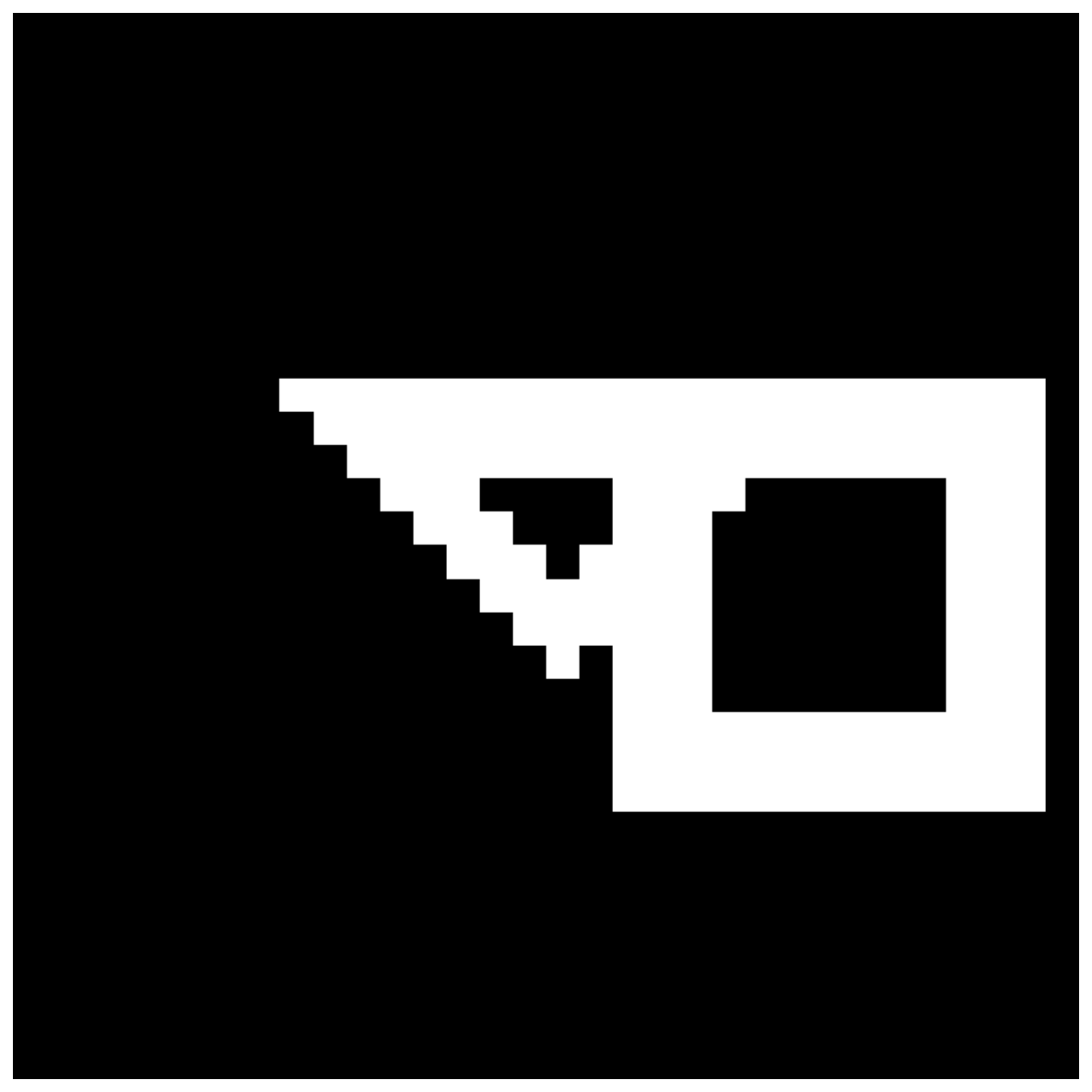}} & 
        \adjustbox{valign=m}{\includegraphics[width=\smallpicsize]{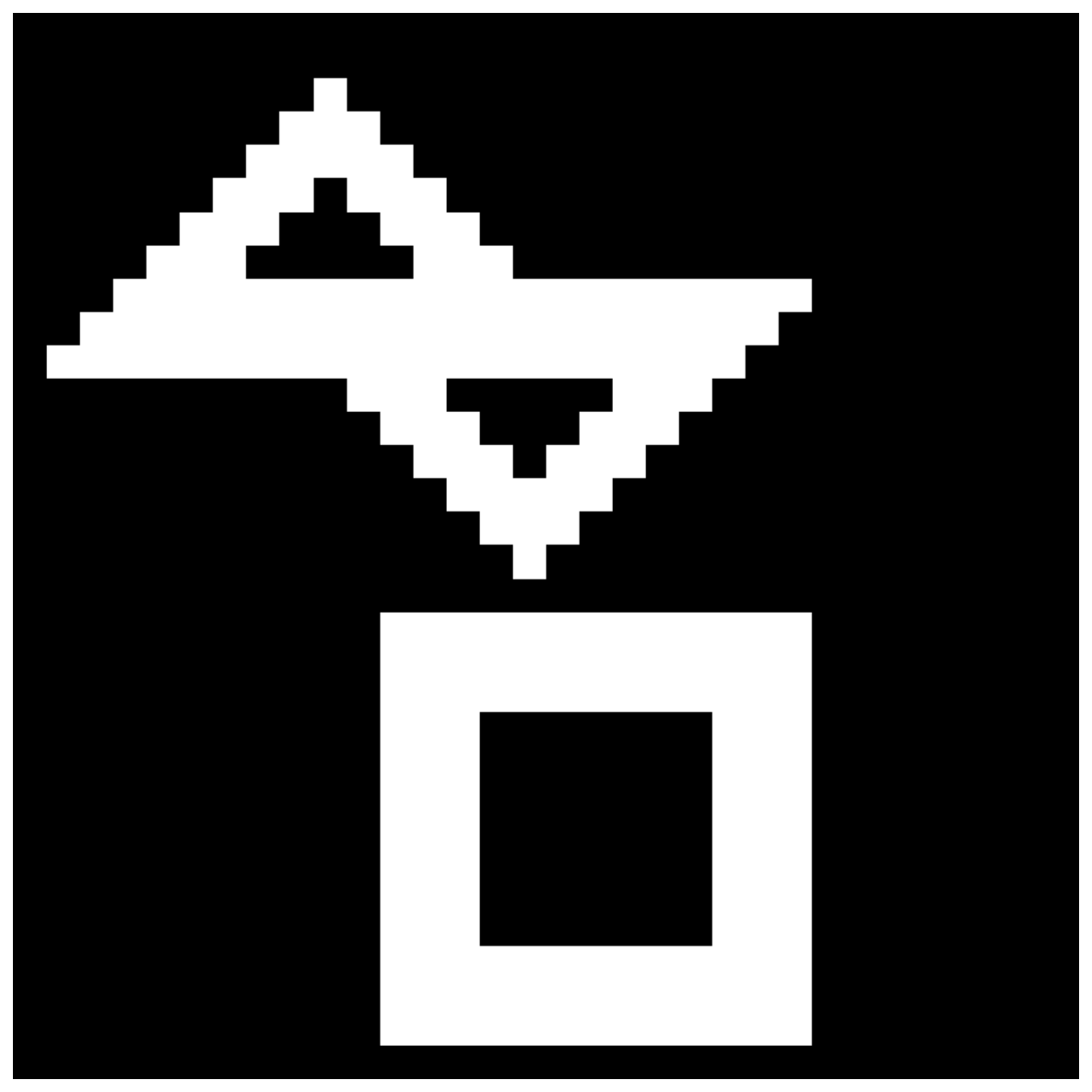}} \\
        \adjustbox{valign=m}{Prediction} & 
        \adjustbox{valign=m}{\includegraphics[width=\smallpicsize]{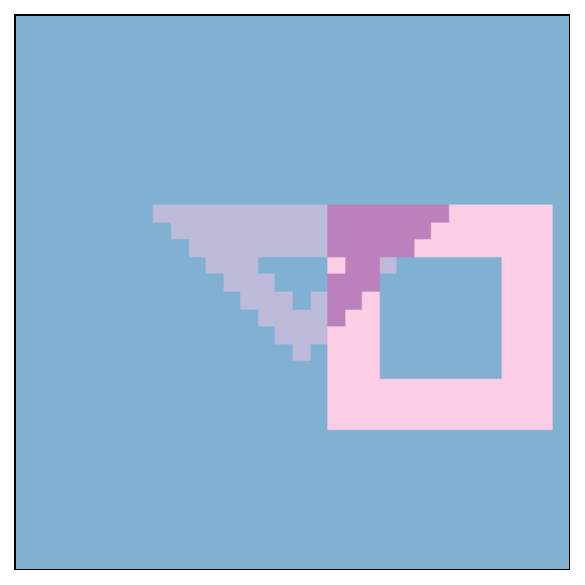}} & 
        \adjustbox{valign=m}{\includegraphics[width=\smallpicsize]{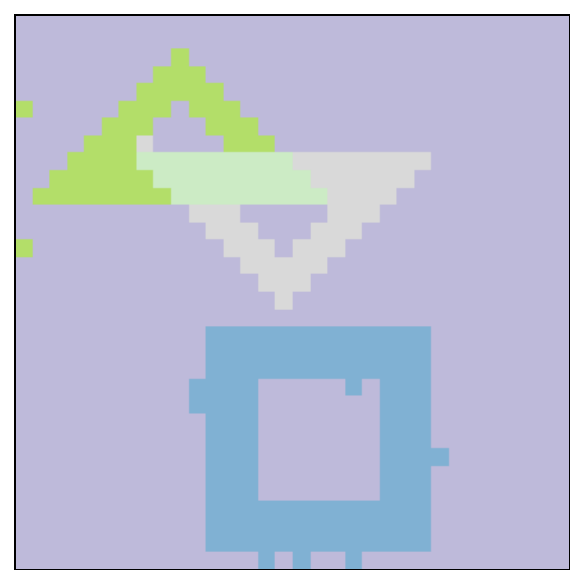}} & 
        \adjustbox{valign=m}{\includegraphics[width=\smallpicsize]{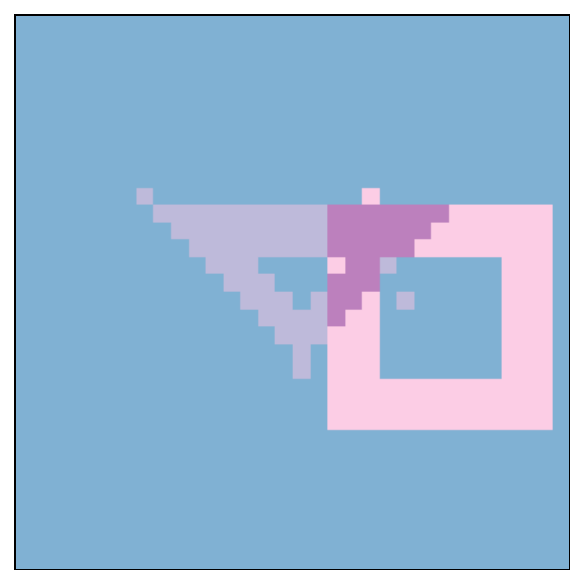}} & 
        \adjustbox{valign=m}{\includegraphics[width=\smallpicsize]{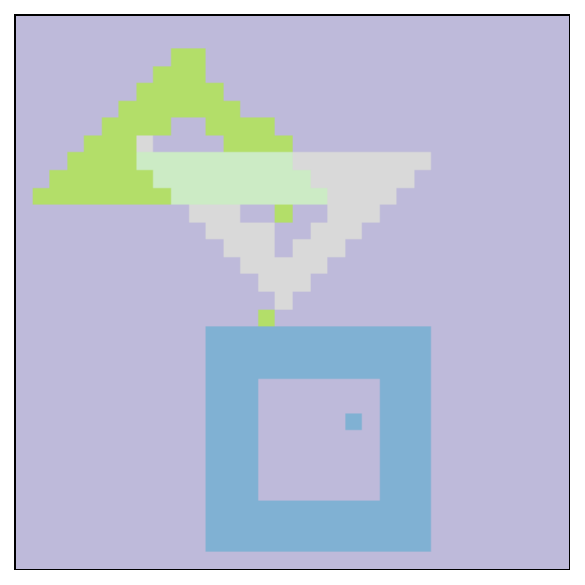}} \\
        \adjustbox{valign=m}{Orientation} & 
        \adjustbox{valign=m}{\includegraphics[width=\smallpicsize]{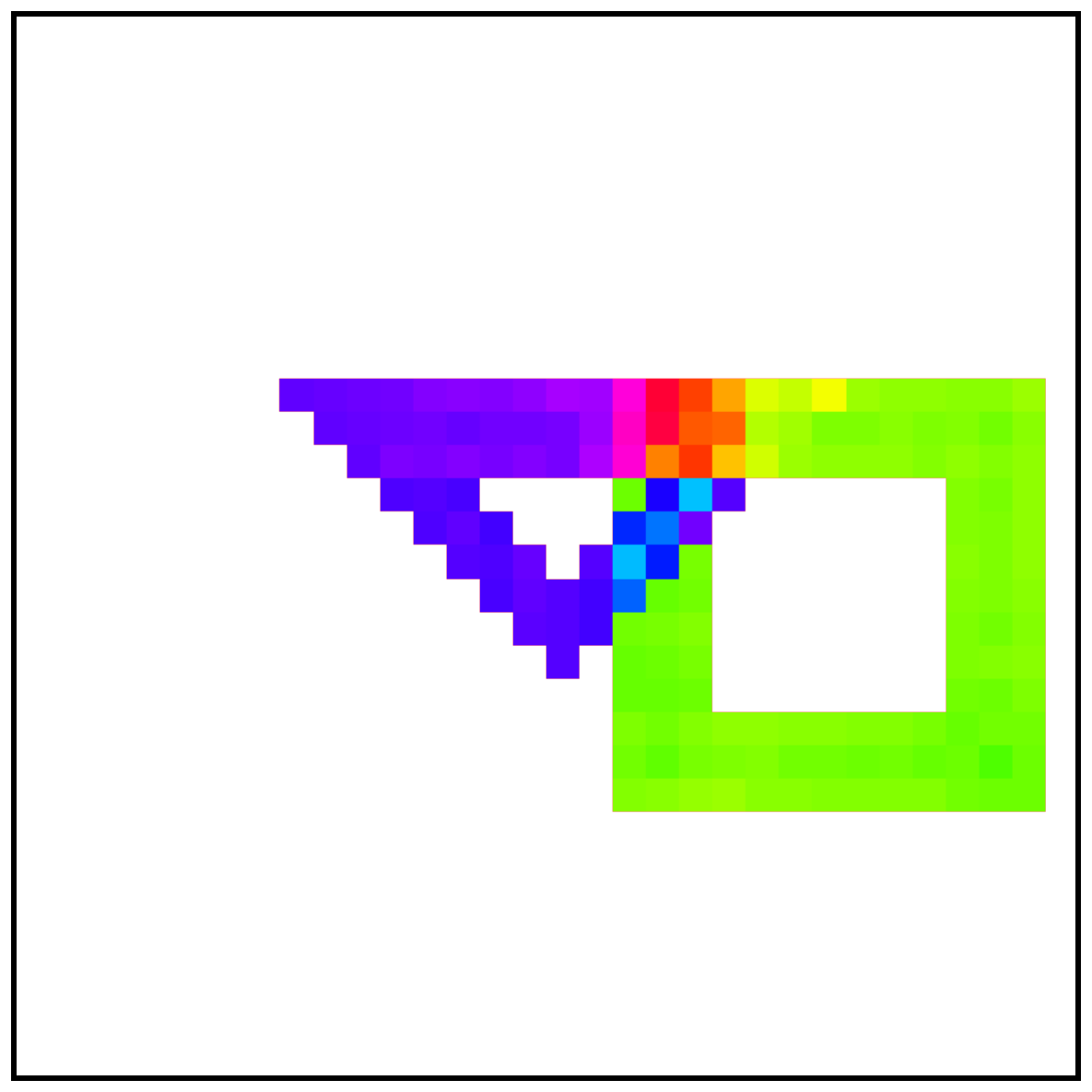}} & 
        \adjustbox{valign=m}{\includegraphics[width=\smallpicsize]{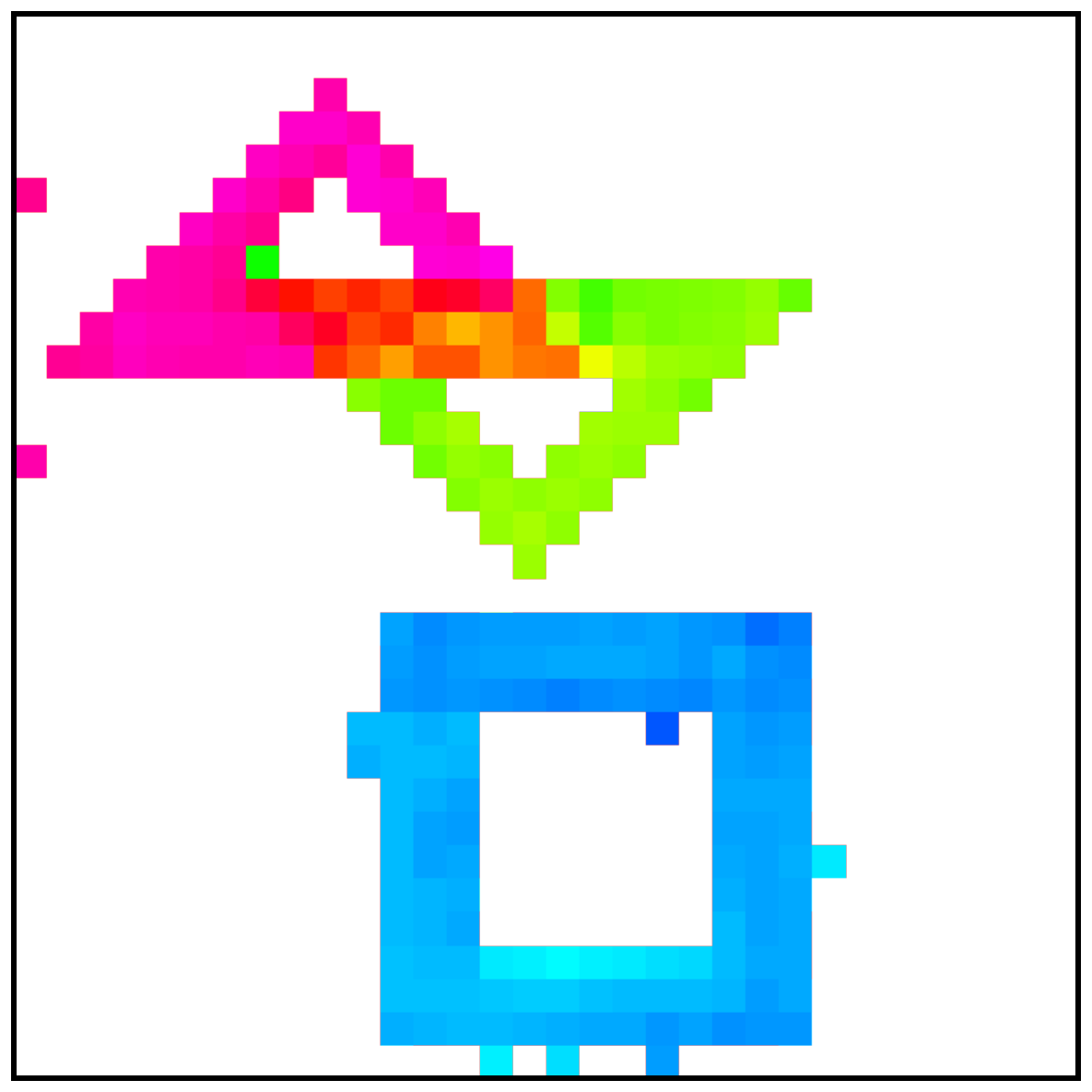}} & 
        \adjustbox{valign=m}{\includegraphics[width=\smallpicsize]{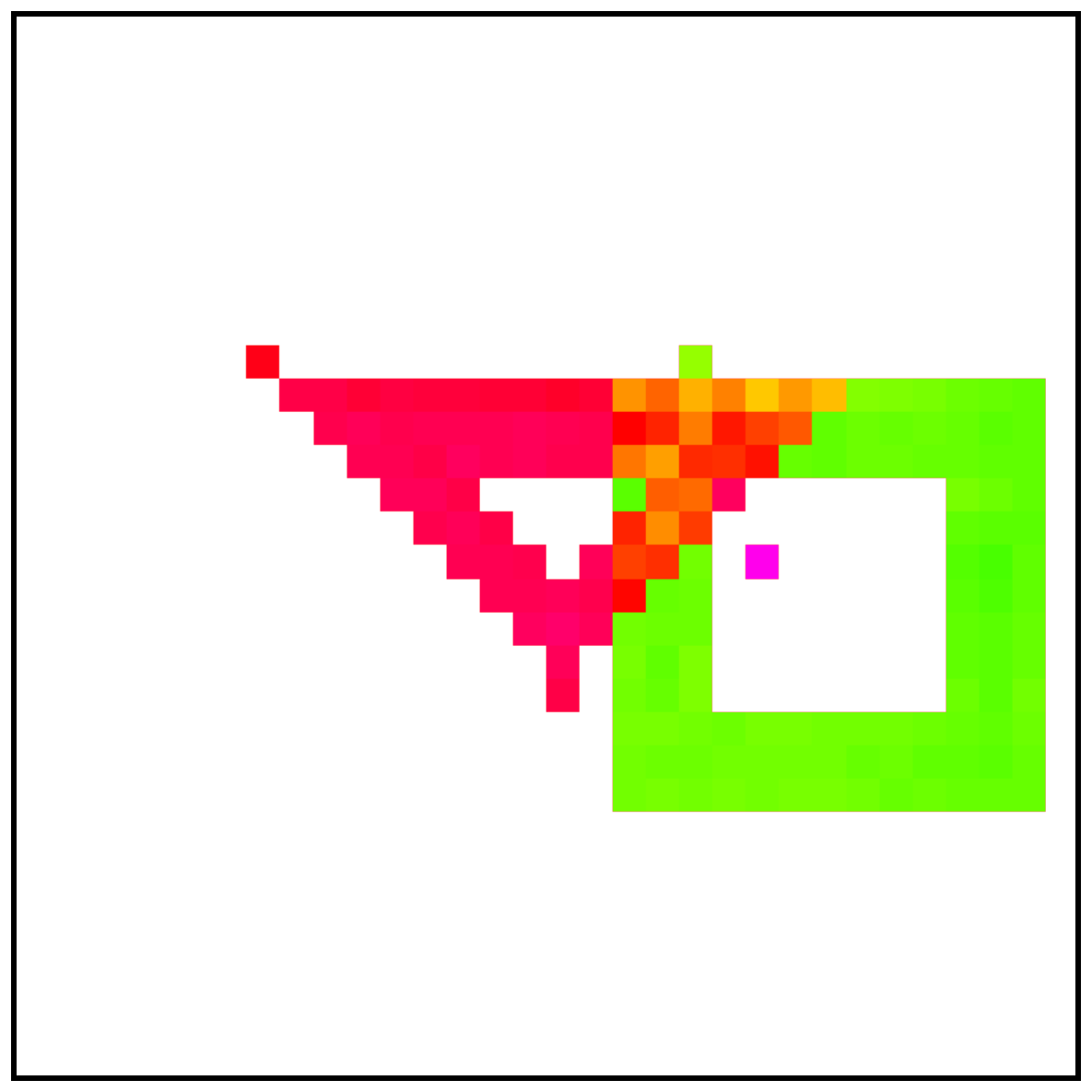}} & 
        \adjustbox{valign=m}{\includegraphics[width=\smallpicsize]{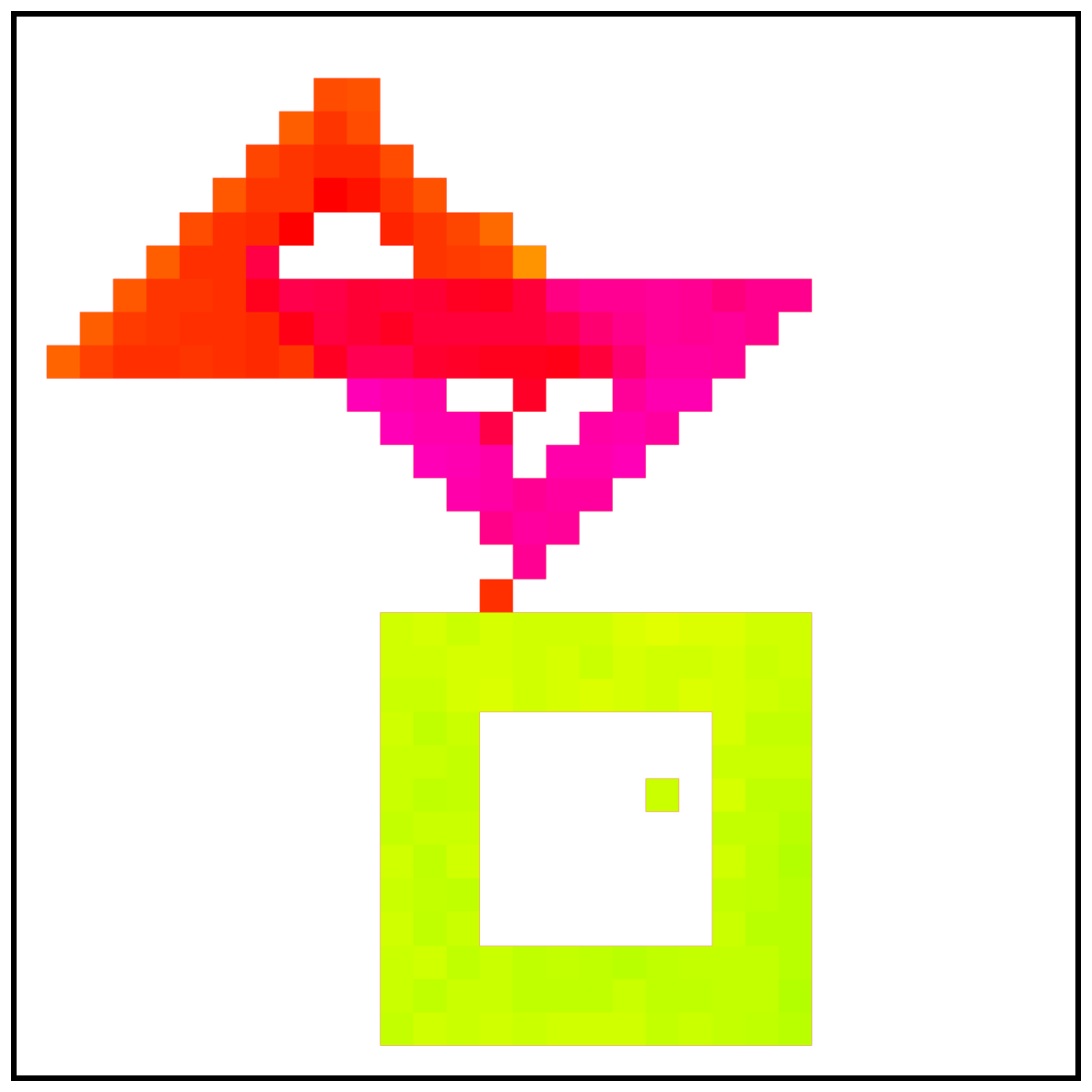}} \\
        &
        \includegraphics[width=\smallpicsize]{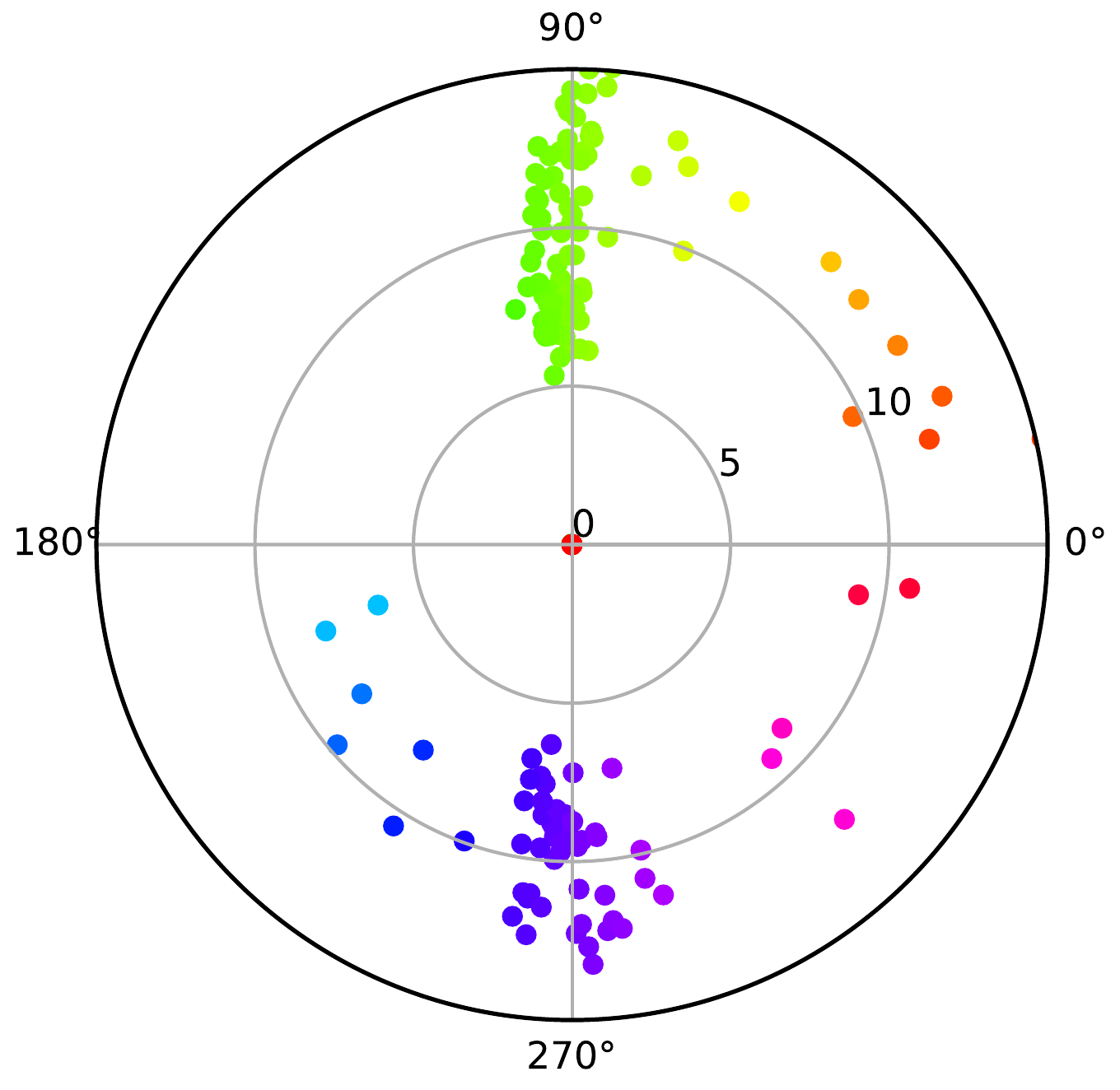} & 
        \includegraphics[width=\smallpicsize]{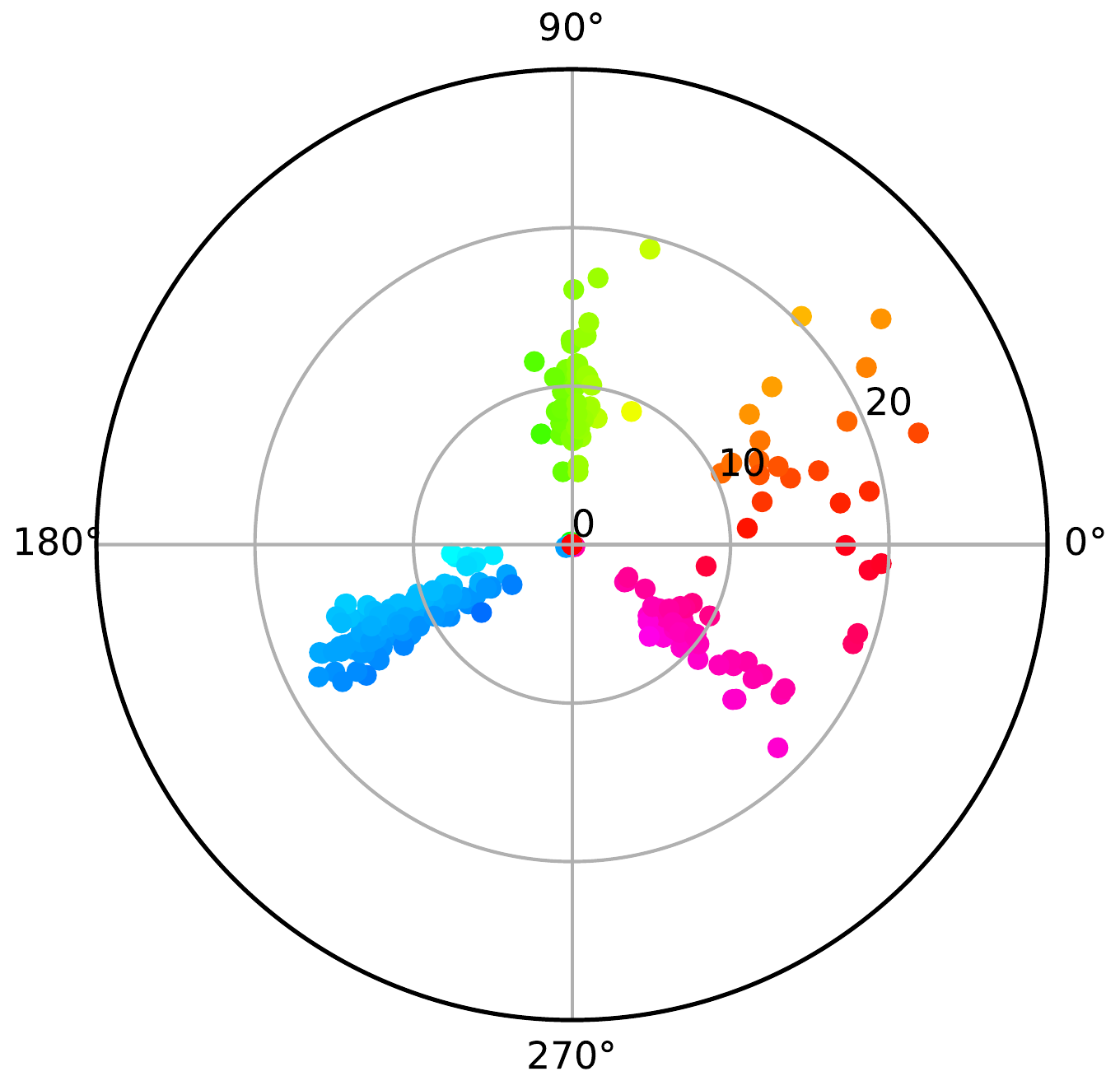} & 
        \includegraphics[width=\smallpicsize]{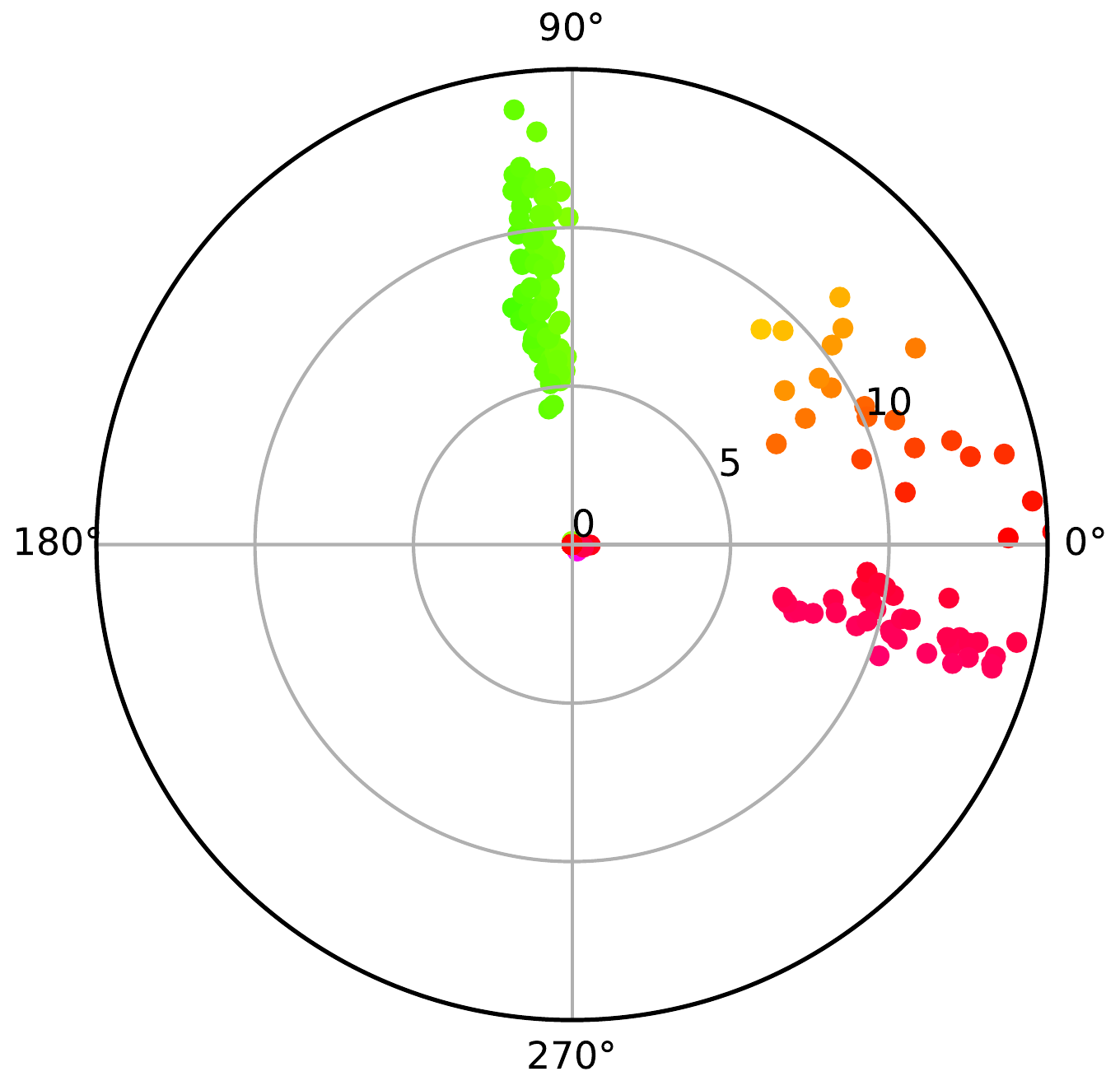} & 
        \includegraphics[width=\smallpicsize]{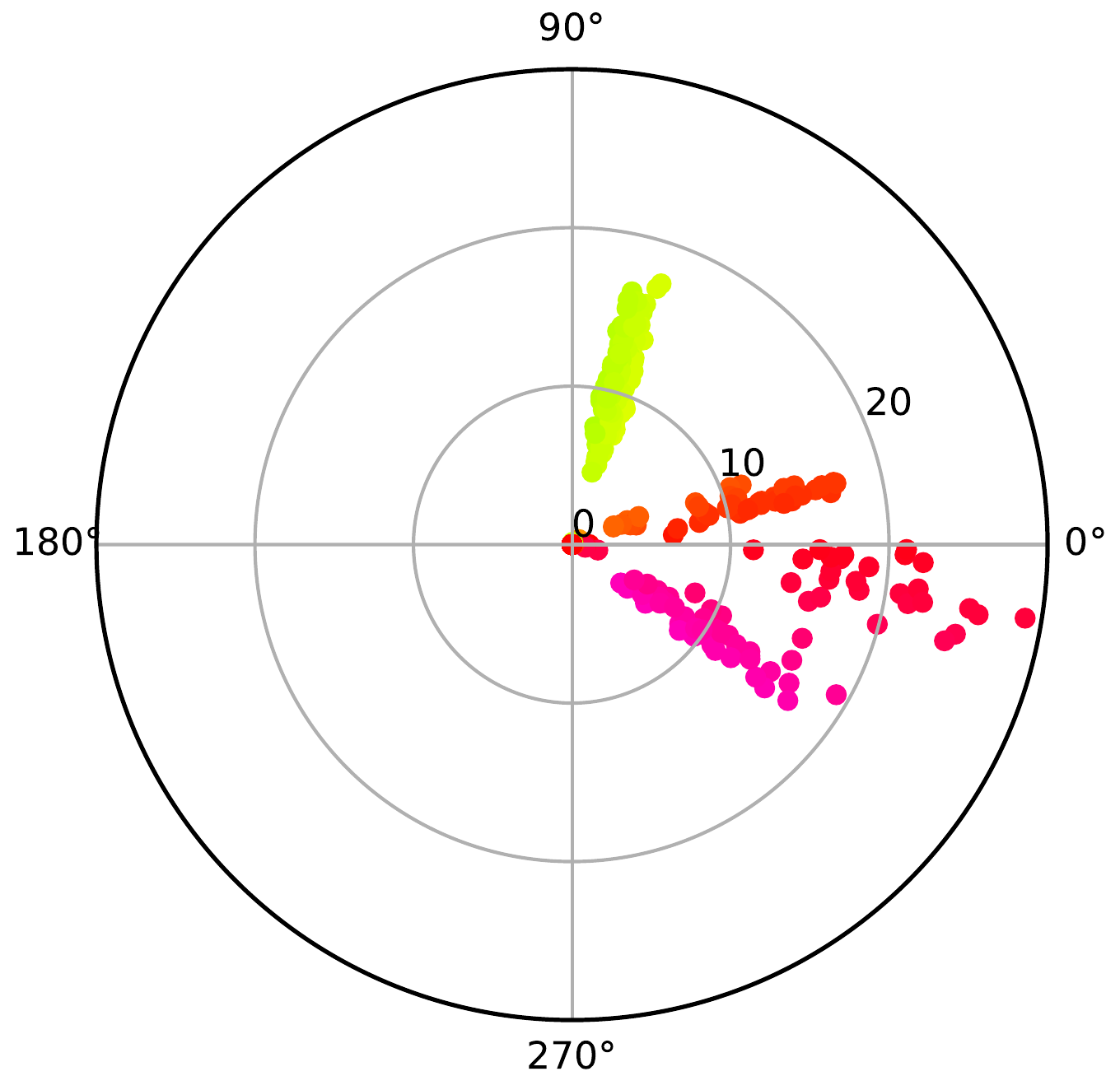} \\
    \end{tabular}
    \caption{The $\gatingout$- and cosine binding mechanisms assign different orientations to objects. With $\gatingout$-binding (\textbf{left}), Rotating Features adopt maximally separated orientations for different objects -- about $180^{\circ}$ apart for two objects, and roughly $120^{\circ}$ apart when three are present. The cosine binding mechanism (\textbf{right}), on the other hand, results in a tighter clustering of orientations for different objects. Nonetheless, both mechanisms create accurate segmentation masks in their predictions.
    }
    \label{fig:objects_on_circle}
\end{figure}

\section{Related Work}
The binding problem in machine learning has been addressed by numerous strategies, primarily focusing on slot-based object-centric representations \citep{greff2020binding}. Here, each ``slot'' is a feature vector within the network that represents one object in a given scene. Methods for differentiating slots include ordering them \citep{eslami2016attend, burgess2019monet, engelcke2019genesis}, spatial coordination \citep{santoro2017simple, crawford2019spatially, anciukevicius2020object, deng2020generative, lin2020space, chen2021roots, du2021unsupervised, baldassarre2022towards}, type-specific slots \citep{hinton2011transforming, hinton2018matrix, von2020towards,wen2022self}, iterative procedures \citep{greff2016tagger, greff2017neural,greff2019multi, stelzner2019faster, kosiorek2020conditional, du2021energy, goyal2021recurrent, zhang2023robust, locatello2020object, lowe2020learning, stelzner2021decomposing, sajjadi2022object, singh2022illiterate, smith2022unsupervised, jia2023unsupervised, jiang2023object, wang2023object, elsayed2022savi, kipf2022conditional, singh2022simple, gopalakrishnan2023unsupervised, wu2023slotformer, seitzer2023bridging,wu2023inverted}, or a combination of these techniques \citep{engelcke2021genesis, wu2022obpose, singh2023neural, traub2023learning}. Some theoretical works have sought to understand these methods' inner workings \citep{lachapelle2023additive} or establish conditions for provably learning slot-based object-centric representations \citep{brady2023provably}. The implications of slot-based object-centric representation on the generalization capabilities of machine learning models have also been studied \citep{dittadi2021generalization}.

Conversely, there has been limited exploration in alternative formats for object representations. \citet{zheng2022dance, zheng2023gust} develop object representations based on the synchronization of spiking neurons. Inspired by the same neuroscientific theories, Rotating Features incorporate rotation dimensions into each feature, allowing for the expression of object affiliation via orientation. Similar representational formats have been explored in supervised and weakly supervised methods \citep{mozer1992learning, zemel1995lending, rao2008unsupervised, rao2010objective, rao2011effects}. In the unsupervised setting, all existing works employ the $\gatingout$-binding mechanism  \citep{reichert2013neuronal, lowe2022complex, stanic2023contrastive, lowe2023rotating}. Our paper marks the first attempt to provide insights on the inner workings of these approaches.

\section{Discussion \& Conclusion}

In conclusion, this paper introduces a novel cosine binding mechanism for Rotating Features that advances our understanding of the binding problem in machine learning. While aligning with biological processes and following an intuitive approach, this mechanism demonstrates a comparable object discovery performance to the existing $\gatingout$-binding mechanism. 
As a result, we learn that Rotating Features require a mechanism that processes features based on their relative orientations. Specifically, when a group of input features collectively generates an output, their contribution to this output is adjusted based on their relative alignment, thereby reducing the influence of misaligned inputs. This ultimately promotes the disentanglement of object-centric representations as features describing a certain object are given the same orientation and therefore processed together, while features from other objects are given divergent orientations to minimize interference.
Essentially, the network can create distinct streams of information by assigning different orientations to features, and learns to utilize these streams to process objects independently. We hope that these insights will inform and inspire further developments in Rotating Features models, ultimately leading to enhanced generalization and reasoning capabilities in machine learning systems.

\subsubsection*{Acknowledgments}
We thank Jascha Sohl-Dickstein and Phillip Lippe for insightful discussions. Sindy Löwe was supported by the Google PhD Fellowship.

\bibliographystyle{iclr2024_conference}
\bibliography{bibliography}

\newpage
\appendix
\section{Appendix}

\subsection{Implementation Details} \label{app:implementation}
Throughout our experiments, we closely follow the implementation details as described by \citet{lowe2023rotating}. Thus, we make use of the same architecture, hyperparameters, dataset processing and evaluation procedure, and only exchange the implementation of the binding mechanism within each layer. Since the cosine binding mechanism has higher memory requirements, we reduce the batch-size to eight on the FoodSeg103 and Pascal VOC datasets, and increase the number of training steps to 240k steps on the FoodSeg103 dataset and to 70k steps on the Pascal VOC dataset.

\subsection{Runtime and Memory Efficiency}\label{app:efficiency}
To evaluate the runtime and memory efficiency of the two binding mechanisms, we applied the Rotating Features models on the 3Shapes dataset \citep{lowe2022complex}. We ensured comparable conditions by running them in the same code environment, where only the layer implementation was exchanged. Additionally, we ran the profiling on the same machine with an Nvidia GTX 1080Ti. To compare the runtime of the two binding mechanisms, we made use of the \texttt{torch.utils.benchmark.Timer} functionality \citep{pytorch}, and ran the forward and backward pass 100 times to average out any variability. To measure the GPU memory requirements, we first reset the PyTorch memory statistics by calling \texttt{torch.cuda.reset\_peak\_memory\_stats()}, then ran the forward and backward pass through our models, before printing the peak memory consumption using \texttt{torch.cuda.max\_memory\_allocated()}.

\end{document}